\documentclass{bmvc2k}

\title{Bird Species Categorization Using Pose Normalized Deep Convolutional Nets}

\addauthor{Steve Branson}{sbranson@caltech.edu}{1}{*}
\addauthor{Grant Van Horn}{gvanhorn@ucsd.edu}{2}{*}
\addauthor{Serge Belongie}{tech.cornell.edu}{3}{}
\addauthor{Pietro Perona}{perona@caltech.edu}{1}{}
\addinstitution{
 California Institute of Technology
 Pasadena, CA, USA
}
\addinstitution{
 University of California, San Diego\\
 La Jolla, CA, USA
}
\addinstitution{
 Cornell Tech\\
 New York, NY, USA
}
\runninghead{Branson, Van Horn et al.}{Bird Species Categorization}

\def\eg{\emph{e.g}\bmvaOneDot}

\usepackage{graphicx,subfigure}
\usepackage{algorithmic,algorithm}
\usepackage{pifont}
\usepackage{multirow}
\usepackage{enumitem}

\newcommand{\Section}[1]{\section{#1}}
\newcommand{\Subsection}[1]{\subsection{#1}}

\newcommand{\Caption}[1]{\caption{#1}}
\newcommand{\paragraphheader}[1]{\par\vspace{0.5mm}\noindent\textbf{#1}}
\newcommand{\squeeze}{\vspace{0mm}}
\newcommand{\unsqueeze}{\vspace{0mm}}

\usepackage[font={color=bmv@captioncolor,footnotesize},labelfont={color=bmv@captioncolor,bf}]{caption}
\newcommand{\coloredparagraphheader}[1]{\par\vspace{0.5mm}\noindent\textbf{\textcolor{bmv@captioncolor}{#1}}}

\DeclareRobustCommand\onedot{\futurelet\@let@token\@onedot}
\def\@onedot{\ifx\@let@token.\else.\null\fi\xspace}

\newcommand{\cmark}{\ding{51}}

\begin{document}

%\showthe\textwidth

\maketitle

\squeeze
\squeeze
\squeeze
\squeeze
\begin{abstract}
We propose an architecture for fine-grained visual categorization that approaches expert human performance in the classification of bird species. Our architecture first computes an estimate of the object's pose; this is used to compute local image features which are, in turn, used for classification. The features are computed by applying deep convolutional nets to image patches that are located and normalized by the pose. We perform an empirical study of a number of pose normalization schemes, including an investigation of higher order geometric warping functions. We propose a novel graph-based clustering algorithm for learning a compact pose normalization space.  We perform a detailed investigation of state-of-the-art deep convolutional feature implementations~\cite{krizhevsky2012imagenet,donahue2013decaf,girshick2013rich,Jia13caffe} and fine-tuning feature learning for fine-grained classification.  We observe that a model that integrates lower-level feature layers with pose-normalized extraction routines and higher-level feature layers with unaligned image features works best.  Our experiments advance state-of-the-art performance on bird species recognition, with a large improvement of correct classification rates over previous methods (75\% vs. 55-65\%). 
\end{abstract}

\squeeze
\squeeze

%\begin{figure}[t]
%\centering
%\includegraphics[width=.45\linewidth]{figures/new/ipad1.png} 
%\Caption{\textbf{TODO: Screen Capture of Demo App}}
%\label{fig:app}
%\end{figure}

\iffalse
\begin{figure}[h!]
  \caption{System Overview}
  \centering
    \includegraphics[width=1.0\textwidth]{figs/System_Figure}
\end{figure}

\begin{figure}[h!]
  \caption{Pipeline}
  \centering
    \includegraphics[width=1.0\textwidth]{figs/PipelineFigure}
\end{figure}
\fi

\squeeze
\Section{Introduction}
\label{sec:introduction}

\squeeze
Fine-grained categorization, also known as subcategory recognition, is a rapidly growing subfield in object recognition.  Applications include distinguishing different types of flowers~\cite{nilsback2006visual,nilsback2008automated}, plants~\cite{kumar2012leafsnap,belhumeur-searching}, insects~\cite{martınez2009dictionary,larios2010haar}, birds~\cite{branson2010visual,farrell2011birdlets,wah2011multiclass,zhang2012pose,bergpoof,zhang2013dpd,chai2013symbiotic,lazebnik2005maximum}, dogs~\cite{parkhi2011truth,liu2012dog,vedaldi2012cats,khoslanovel}, vehicles~\cite{stark2011fine}, shoes~\cite{berg2010automatic}, or architectural styles~\cite{maji2012parts}.  Each of these domains individually is of particular importance to its constituent enthusiasts; moreover, it has been shown that the mistakes of state-of-the-art recognition algorithms on the ImageNet Challenge usually pertain to distinguishing related subcategories~\cite{russakovskydetecting}.  Developing algorithms that perform well within specific fine-grained domains can provide valuable insight into what types of models, representations, learning algorithms, and annotation types might be necessary to solve visual recognition at performance levels that are good enough for practical use.  

\begin{figure}[h!]
  \centering
    \includegraphics[width=1.0\textwidth]{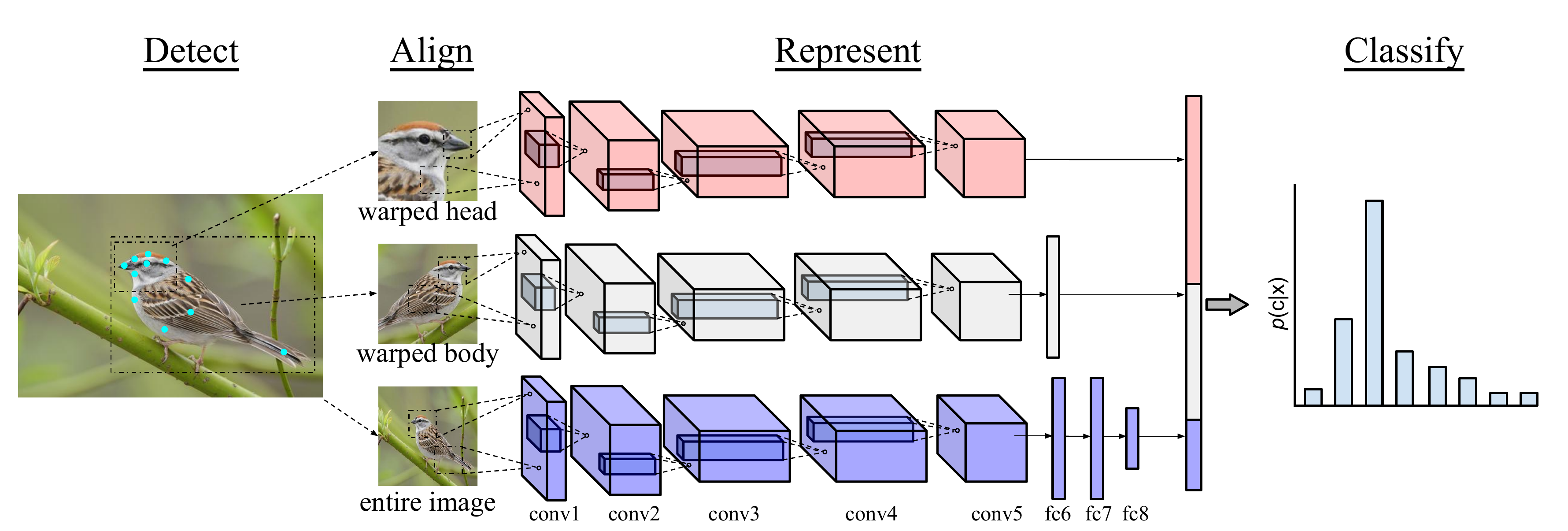}
    \vspace{1mm}
  \caption{\textbf{Pipeline Overview:} Given a test image, we use groups of detected keypoints to compute multiple warped image regions that are aligned with prototypical models.  Each region is fed through a deep convolutional network, and features are extracted from multiple layers.  Features are concatenated and fed to  a classifier.}
\end{figure}

Within fine-grained categorization, bird species recognition has emerged as one of the most widely studied areas (if not the most) in the last few years, in part due to the release of CUB-200~\cite{welinder2010caltech} and CUB-200-2011~\cite{WahCUB_200_2011} as standard datasets.  Performance improvements on the CUB datasets over the last few years have been remarkable, with early methods achieving $10-20\%$ 200-way classification accuracy~\cite{branson2010visual,WahCUB_200_2011,welinder2010caltech,yao2011combining}, and recent methods achieving $55-65\%$ accuracy~\cite{bergpoof,gaves2013fgvc,chai2013symbiotic,zhang2013dpd,branson2014ignorant,donahue2013decaf}.  Here we report further accuracy gains up to $75.7\%$.  
%Machine-based bird species recognition has reached the level where it is good enough to be useful for practical applications~\cite{berg-birdsnap-cvpr2014,branson2014ignorant}.  Our improvements were greatly facilitated by the recent success of new deep convolutional neural network features~\cite{krizhevsky2012imagenet,donahue2013decaf,girshick2013rich,Jia13caffe}, and we study their integration with part-level alignment models and adaptation to fine-grained problems.  
This paper makes 2 main contributions: 
\begin{enumerate}[noitemsep,nolistsep]
\itemsep0em
\item An empirical study of pose normalization schemes for fine-grained classification, including an investigation of higher order geometric warping functions and a novel graph-based clustering algorithm for learning a compact pose normalization space. 
\item A detailed experimental investigation of state-of-the-art deep convolutional features~\cite{krizhevsky2012imagenet,donahue2013decaf,girshick2013rich,Jia13caffe} and feature learning for fine-grained classification.  %We observe that a model that integrates lower-level features with pose-normalized localization, and higher-level features with unaligned image features works best.
%\item Results significantly outperforming state-of-the-art performance on CUB-200-2011, and detailed studies of the main factors that led to good performance. 
\end{enumerate}
\vspace{2mm}

%Birds have a large and devoted following--with an estimated 46.7 million Americans involved in observing or photgraphing birds in 2011~\cite{fish2011}.  

%\paragraphheader{Paper structure:} The structure of the paper is as follows....

\Section{Related Work}
\label{sec:related_work}

Work on fine-grained categorization over the past 5 years has been extensive.  Areas explored include feature representations that better preserve fine-grained information~\cite{yao2011combining,CVPR12_0654,martınez2009dictionary,yang2012unsupervised}, segmentation-based approaches \cite{nilsback2008automated,chai2011bicos,chai2012tricos,chai2013symbiotic,angelova2013efficient,gaves2013fgvc} that facilitate extraction of purer features, and part/pose normalized feature spaces~\cite{farrell2011birdlets,bergpoof,berg-birdsnap-cvpr2014,wah2011multiclass,parkhi2011truth,zhang2012pose,zhang2013dpd,liu2012dog,vedaldi2012cats}.  Among this large body of work, it is a goal of our paper to empirically investigate which methods and techniques are most important toward achieving good performance.  Consequently, we describe a simple pose normalization method that can be used to express many of the above techniques and logical extensions.  We find that a similarity based pose warping function as used by Berg and Belhumeur \cite{bergpoof} yields the best performance and can be improved by using more parts to estimate the warping, while being made more compact and efficient by learning pose regions.  We investigate the interplay between pose-normalized images and the types of features that work best with them.

The impressive performance of deep convolutional networks~\cite{lecun1995convolutional} (CNNs) on large scale visual recognition challenges, ignited by~\cite{krizhevsky2012imagenet}, has motivated researchers to adapt CNNs that were pre-trained on ImageNet to other domains and datasets, including Caltech-101~\cite{zeiler2013vis}, Caltech-256~\cite{zeiler2013vis}, VOC detection~\cite{girshick2013rich}, and VOC classification~\cite{zeiler2013vis}.  Donahue et al.~\cite{donahue2013decaf} extracted CNN features from part regions detected using a DPM, obtaining state-of-the-art results in bird species classification.  Our work is inspired by these results, and we improve on them by combining ideas inspired from fine-grained recognition and CNN research.  In particular, we find that different layers of the CNN are appropriate for different levels of alignment.  Secondly, we explore different methods for fine-tuning CNN weights on the CUB-200-2011 training set, inspired by techniques and results from Girshick et al.~\cite{girshick2013rich}.

\Section{Pose Normalization Schemes}
\label{sec:pose_norm}

In this section, we define a class of pose normalization schemes based on aligning detected keypoints to the corresponding keypoints in a prototype image.  In Section~\ref{sec:pose_learn}, we introduce an algorithm for learning a set of prototypes that minimizes the pixel-wise alignment error of keypoint annotations in a training set and works for arbitrary warping functions.

\Subsection{Pose Normalization By Prototypical Regions}

Let $\{(X_i,Y_i)\}_{i=1}^n$ be a training set of $n$ images and ground truth part annotations, where each annotation $Y_i=\{y_{ij}\}_{j=1}^K$ labels the pixel location and visibility of $K$ 2D keypoints in the image $X_i$.  Due to its simplicity and ease of collection, this style of 2D keypoint annotations is widely used (\eg, for birds~\cite{WahCUB_200_2011}, dogs~\cite{liu2012dog}, faces~\cite{huang2007labeled}, and humans~\cite{bourdev2009poselets}).  

Let $\Psi(X,Y)=[\psi_p(X,Y)]_{p=1}^P$ be a feature vector that is obtained by concatenating $P$ pose normalized feature spaces, where each $\psi_p(X,Y)$ may correspond to a different part or region of an object and can be estimated using some subset of keypoints in $Y$.  We consider a simple definition of $\psi_p(X,Y)$ based on prototypical examples.  Let the $p$-th prototype $R_p=\{i_p,b_p,S_p\}$ consist of a reference image $i_p$, a rectangle $b_p$ defining a region of interest in $X_{i_p}$ for feature extraction, and a set of keypoint indices $S_p$.  Given a test image $X_t$ with detected keypoints $Y_t$, we solve for the transformation $W(y_{tj},w)$ in some class of warping functions $\mathcal{W}$ that best aligns the corresponding keypoints in $Y_t$ to $Y_{i_p}$:  
\begin{eqnarray}
w_{tp}^*=\arg\min_{w \in \mathcal{W}} \sum_{j \in S_p} E(y_{tj},R_p,w),\ \ \ \ \mathrm{where} \ E(y_{tj},R_p,w) = \| \hat{y}_{i_pj} - W(y_{tj},w) \|^2
\label{eq:warp}
\end{eqnarray}
where $\|\cdot\|$ indicates Euclidean distance, and $\hat{y}_{i_pj}$ is a version of $y_{i_pj}$ after normalizing by the bounding box $b_p$ (by subtracting the upper-left coordinate and dividing by the width/height).  The induced pose normalized feature space $\psi_p(X,Y)=\phi(X(w_{tp}^*))$ is obtained by applying this warp to the image $X_t$ and then extracting some base feature $\phi(X)$, where $X(w)$ is a warped version of image $X$.  %Due to normalizing and cropping by the bounding box $b_p$, $X(w)$ is always the same size.   %In our case, $\phi(X)$ is the output of some layer(s) of a CNN and $X(w)$ is a $256 \times 256$ image.  The decision to define pose regions in terms of prototypical examples is primarily for visualization convenience; it makes it easy and intuitive for a domain expert to define regions of interests and prototypes by hand; examples can be seen in Fig.~\ref{fig:warped_regions}.  However, we will also show in the next section that this definition lends itself well to an algorithm for learning prototypes automatically using a pairwise graph clustering algorithm.

In Table~\ref{tab:warp}, we define how Eq~\ref{eq:warp} can be computed for many different warping families, including simple translations, similarity transformations (2D rotation, scale, and translation), and affine transformations.  The extension to other families such as homographies and thin-plate-splines~\cite{tps,shape_context} is straightforward.  The above transformations have simple closed form solutions and have well understood properties in approximating projective geometry of 3D objects.  In each case, the applicable warping function is only well-defined if the number of points available $|S|$ is sufficiently large.  Let $S \subseteq S_p$ be the subset of points in $S_p$ that are visible as determined by detected keypoints $Y_t$.  If $|S|$ falls below the applicable minimum threshold, we set the induced feature vector $\psi_p(X,Y)$ to zero.

\begin{table}
\footnotesize
\begin{center}
\begin{tabular}{cccc}
Name & $W(y,w)$ & Solve $w_{tp}^*$ & \# Pts\\
\hline
Translation & $y=y_t + T$ & $T=\mu_i-\mu_t$ & $|S| \ge 1$\\
2D Similarity & $y=s R y_t + t$ & $\begin{aligned}R &= V \mathrm{diag}(1,\mathrm{det}(V U^\top)) U^\top,\ \ s = \frac{\mathrm{tr}(\bar{M}_i^\top R \bar{M}_t)}{\mathrm{tr}(\bar{M}_t^\top\bar{M}_t)},\ \ T =\mu_i-sR\mu_t\end{aligned}$ & $|S| \ge 2$\\
2D Affine & $y=A y^h_t$ & $A=M_i M^{h\top}_t (M^h_t M^{h\top}_t)^{-1}$ & $|S| \ge 3$\\
%Homography & $y^h=H y^h_t$ & 4 point algorithm~\cite{fourpoint} & $|S| \ge 4$\\
%Thin-plate-spline & $y=\sum_{j} c_j \varphi(\|y_{tj}-y_{ij}\|) $ & solve for $c_i$ as in~\cite{tps} & $S = S_p$\\
\hline
\end{tabular}
\end{center}
\caption{Computation of warping function $W(y,w)$ from detected points $Y_t[S]$ to a prototype $Y_i[S]$ for different warping families.  In the above notation, let $M_t$ and $M_i$ be $2 \times |S|$ matrices obtained by stacking points in $Y_t$ and $Y_i$, and $\mu_t$ and $\mu_i$ be their means.  Let $\bar{M}_t$ and $\bar{M}_i$ denote mean subtracted versions of these matrices, and the superscript $h$ denote points in homogeneous coordinates.  Let $C=U \Sigma V^\top$ be the SVD of $C= \bar{M}_t \bar{M}_i^\top$.}
\label{tab:warp}
\end{table}

\iffalse
Most popular pose normalization schemes for fine-grained classification can be represented in this schema.  One of the most popular schemes ...
\fi

\begin{figure}[t]
\centering
\includegraphics[width=\linewidth]{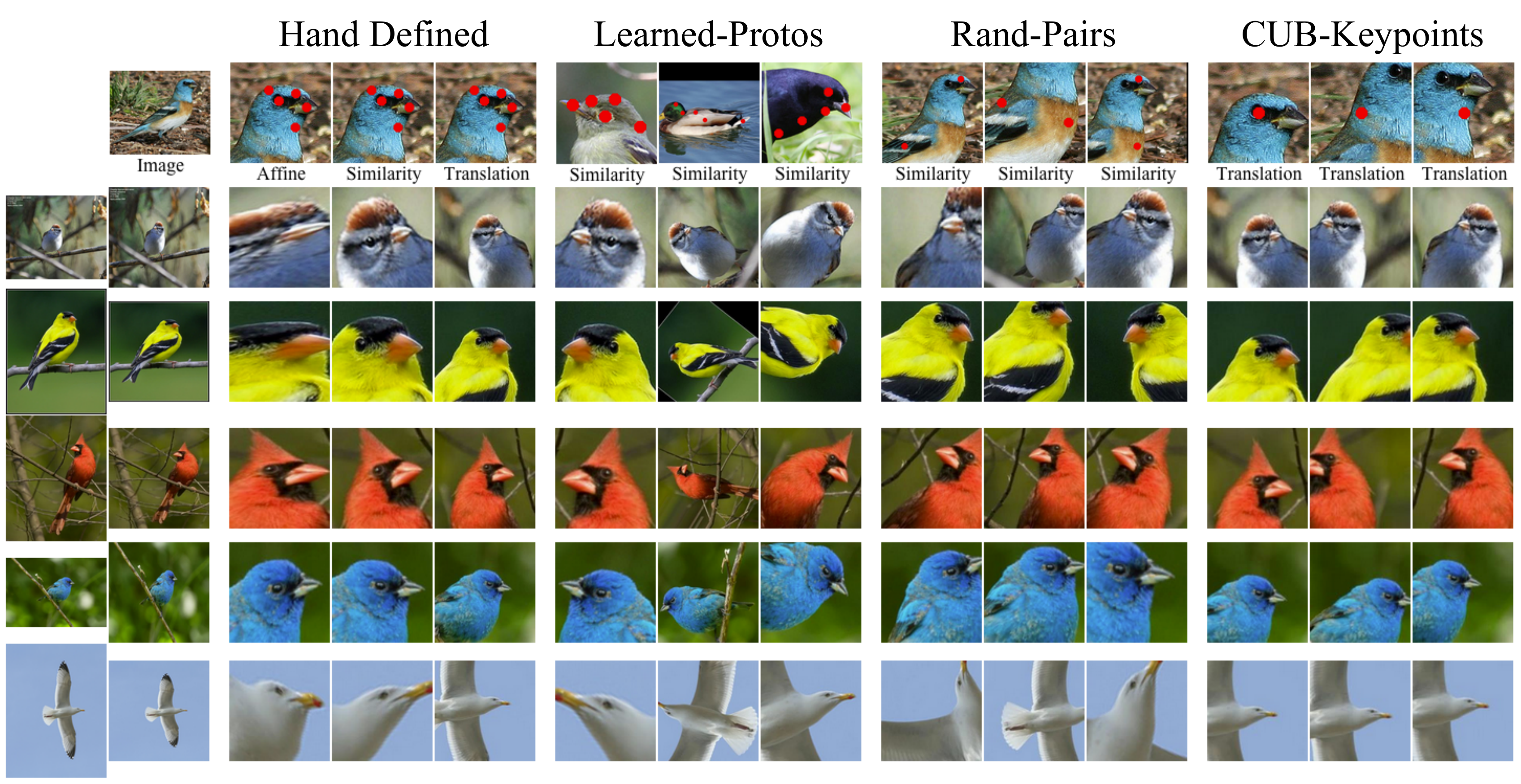}
\unsqueeze
\unsqueeze
\Caption{{\bf Example Warped Regions:} The top row visualizes different prototypes, each of which defines a region of interest and multiple keypoints that are used to estimate a warping.  The bottom rows show the resulting warped regions $X(w_{tp}^*)$ when 5 images are aligned with each prototype.   The 4 groupings of warped regions represent 4 baseline experiments analyzed in Table~\ref{tab:combine}, which includes 1) Hand-Defined head or body regions, 2) the 1st 3 prototypes learned using our method from Section~\ref{sec:pose_learn}, 3) Rand-Pairs, which simulates~\cite{bergpoof}, 4) CUB-Keypoints, which simulates~\cite{branson2014ignorant}.  In general, we see that a similarity transform captures the scale/orientation of an object better than a translation, while an affine transformation sometimes overly distorts the image.  Using more points to estimate the warping allows for non-visible keypoints and ambiguous image flipping problems to be handled consistently.}
\label{fig:warped_regions}
\end{figure}

\iffalse
figure -- show examples for different alignment models

figure -- show part detection
\fi

\Subsection{Learning Pose Prototypes}
\label{sec:pose_learn}

In this section, we introduce an algorithm for learning pose prototypes from training images with keypoint annotations.  The approach has a similar objective to a poselet learning algorithm~\cite{bourdev2009poselets}.  The main difference is that our approach generalizes to arbitrary warping schemes while explicitly optimizing pixel-wise alignment error in the induced feature space.

Recall that $E(y_{tj}, R_p, w^*_{tp})$ is the squared pixel alignment error in the induced pose-normalized feature space when matching image $t$ to prototype $R_p$.  We attempt to learn a set of prototypes $\{R_p\}_{p=1}^P$ that minimizes the alignment error under the constraint that each keypoint $y_{tj}$ in the training set must be aligned with low error to at least one prototype.  The intuitive justification is that all portions of an object -- each of which might contain important discriminative information -- should be consistently aligned in at least one component of the feature space $\Psi(X,Y)=[\psi_p(X,Y)]_{p=1}^P$.  Our goal is to choose a set of prototypes $\mathbf{R}^*$ that optimizes the objective
\begin{equation}
\mathbf{R}^*=\arg\min_{\mathbf{R}} \lambda P + \frac{1}{nK} \sum_{t=1}^n \sum_{j=1}^K \min_{p} E(y_{tj}, R_p, w^*_{tp})
\label{eq:learn_proto}
\end{equation}
where the first term penalizes the number of prototypes selected and the second term minimizes pixel-wise alignment error, with $\lambda$ being a tweakable tradeoff parameter.  The optimization problem can be complex due to the possibility of complex warping functions $w^*_{tp}$ (Eq~\ref{eq:warp}) and prototype definitions.  To make the problem tractible, we consider candidate prototypes anchored by a keypoint in the training set ($nK$ candidates in total).  Given an anchor $y_{ik}$ representing the $k$-th keypoint in image $i$, we define the candidate prototype $R_{ik}=\{i,S_{ik},b_{ik}\}$ in terms of a set $S_{ik}$ of the $M$-nearest neighbors in $Y_i$ to $y_{ik}$, and $b_{ik}$ as an expanded bounding box around those keypoints. 

We can solve Eq.~\ref{eq:learn_proto} as a non-metric facility location problem~\cite{erlenkotter1978dual}. Given predefined costs $c_{lm}$ of connecting city $l$ to facility $m$ and costs $o_m$ of opening facilities, the goal is to open a subset of facilities with minimum cost such that each city must be connected to one facility.  Eq~\ref{eq:learn_proto} reduces to a facility location problem, where each anchor point $y_{ik}$ is a candidate facility with open cost $\lambda$, and each keypoint $y_{tj}$ is a city that can be connected to a facilty with cost $c_{tj,ik}=E(y_{tj}, R_{ik}, w^*_{t,ik})$.  A nice property of facility location problems is that, unlike some clustering algorithms like k-means, a fast greedy algorithm~\cite{jain2003greedy} has good approximation guarantees ($1.61$~\cite{jain2003greedy} when the city-facility costs are metric, $1+\mathrm{ln} P$~\cite{hochbaum1982heuristics} for non-metric costs).  This algorithm requires precomputing pairwise costs $c_{tj,ik}$ and sorting them.  Examples of learned prototypes can be seen in Fig~\ref{fig:warped_regions}.

%-------------------------------------------------------------------------
\Section{Deep Convolutional Features}

\label{sec:deep}

Our pose-warped image regions $\{ X(w_{tp}^*) \}_p$ are each fed into a feature extractor $\phi(X)$, where $\phi(X)$ is the output of one or more layers of a deep convolutional neural network (CNN)~\cite{krizhevsky2012imagenet}.  We use the network structure from Krizhevsky et al.~\cite{krizhevsky2012imagenet}.

%In brief, a fixed input image region of size 224x224 is forward propagated through 5 convolutional layers, followed by 3 fully connected layers. The last fully connected layer outputs data to an N-way softmax function, where N is the number of classes.  Whereas a rush of papers have begun to adapt CNNs to different computer vision problems with very compelling results~\cite{zeiler2013vis,donahue2013decaf,girshick2013rich,zhang2013panda}, many aspects of their performance and training are still not well understood.  In the next two sections, we provide an empirical investigation of a couple of different aspects of CNN features that to our knoweldge have not been extensively studied yet.

\Subsection{Multi-Layer CNN Features For Different Alignment Models}

The progression through the 8-layer CNN network can be thought of as a progression from low to mid to high-level features.  The later layers aggregate more complex structural information across larger scales--sequences of convolutional layers interleaved with max-pooling are capable of capturing deformable parts, and fully connected layers can capture complex co-occurence statistics.  On the other hand, later layers preserve less and less spatial information, as max-pooling between each convolutional layer successively reduces the resolution of the convolutional output, and fully connected layers drop semantics of spatial location.  We thus hypothesize (and verify empirically in Section~\ref{sec:experiments}), that different layers of this pipeline are more appropriate for different alignment models, and combining multiple levels of alignment can yield superior performance.  %For example, an image-level alignment scheme, where features are extracted from the entire image, will work best with the last fully connected layers (as empirically verified in~\cite{krizhevsky2012imagenet}).  On the other hand, a pose-aligned region may work better with the earlier convolutional layers.  Furthermore, combining multiple levels of alignment can yield superior performance.  Image-level features include greater contextual information and remain unchanged when part detection is noisy.  By contrast, pose-aligned features preserve finer-grained discriminative information but only work if part detection succeeds.  %Let $R_p$ be a region as defined in Section~\ref{sec:pose_norm}, and assume we associate $R_p$ with a feature layer $l_p$.  We denote the corresponding feature vectors $\phi_{l_p}(X(w_{tp}^*))$ as the output of the $l_p$-th layer of the CNN given the appropriate pose-warped image $X(w_{tp}^*)$ as an input.

Our final feature space concatenates features from multiple regions and layers, and one-vs-all linear SVMs are used to learn weights on each feature.  The use of an SVM (instead of the multiclass logistic loss used by CNNs) is primarily for technical convenience when combining multiple regions.  To handle layers with different scales of magnitude, each CNN layer output is normalized independently during feature extraction.  In the next section, we explore a few different approaches for training the internal weights of the CNN.

\Subsection{Training the Convolutional Neural Net}
\label{sec:cnn_train}

We consider 4 training/initialization methods:

\paragraphheader{Pre-Trained ImageNet Model:} This corresponds to the methodology explored in~\cite{donahue2013decaf}, where the CNN is pre-trained on the 1.2 million image ImageNet dataset and used directly as a feature extractor.  

\paragraphheader{Fine-Tuning the ImageNet Model:} This corresponds to the methodology explored in~\cite{girshick2013rich}. Here, the final 1000-class ImageNet output layer is chopped off and replaced by a 200-class CUB-200-2011 output layer.  The weights of the new layer are initialized randomly, and stochastic gradient descent (SGD) and back propagation are used to train the entire network jointly with a small learning rate.  Because the last layer is new and its weights are random, its weights are likely much further from convergence than the pre-trained ImageNet layers.   Consequently, its learning is increased by a factor of 10. 

\paragraphheader{Two Step Fine-Tuning Method:} We explore a 2nd possible fine-tuning method that aims to avoid using unbalanced learning rates for different layers.  Here, we use the same network structure as for the previous method.  We use a two step process.  In the first step, we fix the weights of the old ImageNet layers and learn the weights of the new 200-class output layer--this is equivalent to training a multiclass logistic regression model using the pre-trained ImageNet model as a feature extractor.  It is a fast, convex optimization problem.  This fixes the problem of initializing the new layer.  SGD and back propagation are then used to jointly train all weights of the entire network, where each layer is given the same learning rate.  To our knowledge, this initialization scheme has not yet been explored in earlier work.

\paragraphheader{Training From Scratch:} The earlier three approaches can be seen as an application of transfer learning, where information from the ImageNet dataset has been used to train a better classifier on a different set of classes/images.  To help differentiate between gains from more training data and the network structure of the CNN, we investigate training the CNN without ImageNet initialization.  Weights are initialized randomly before training with SGD.  %We note that since the CNN model has 60 million learnable parameters~\cite{krizhevsky2012imagenet} and the CUB-200-2011 dataset has $<6000$ training images, overfitting is a major concern.  Nevertheless, we observe that training the internal CNN parameters is effective in practice (Section~\ref{sec:experiments_train}) and the choice of initialization strategy is important.

\Section{Experiments}

\label{sec:experiments}

We evaluate performance on the CUB-200-2011 dataset~\cite{WahCUB_200_2011}, a challenging dataset of 200 bird species and $11,788$ images. The dataset includes annotations of 15 semantic keypoint locations.  We use the standard train/test split and report results in terms of classification accuracy.  Although we believe our methods will generalize to other fine-grained datasets, we forgo experiments on other datasets in favor of performing more extensive empirical studies and analysis of the most important factors to achieving good performance on CUB-200-2011.  Specifically, we analyze the effect of different types of features, alignment models, and CNN learning methods.  We believe that the results of these experiments will be informative and useful to researchers who work on object recognition in general.

\coloredparagraphheader{Implementation Details:} We used the DPM implementation from~\cite{branson2013efficient}, which outputs predicted 2D locations and visibility of 13 semantic part keypoints.  To learn pose prototype regions, we chose $\lambda=8^2$, which means that a new prototype should be added if it reduces the average keypoint alignment error by 8 pixels.  For our best classifier, we concatenated features extracted from each prototype region with features extracted from the entire image.

We used the Caffe code base from Jia~\cite{Jia13caffe} to extract, train, and fine-tune the CNN with the default structure and parameter settings.  When extracting feature outputs from different CNN layers, we use the names \textit{conv3}, \textit{conv4}, \textit{conv5}, \textit{fc6}, and \textit{fc7}, where \textit{conv} denotes a convolutional layer, \textit{fc} denotes a fully connected layer, and the number indicates the layer number in the full CNN.  We appended these names with the suffix \textit{-ft} to denote features extracted on a CNN that was fine-tuned on CUB-200-2011.  To fine-tune the CNN, we set the base learning rate to $0.001$.

\Subsection{Summary of Results and Comparison to Related Work}

\begin{table}[ht]
\centering
\footnotesize
\begin{tabular}{lcccccc}
\hline
Method & Oracle Parts & Oracle BBox & Part Scheme & Features & Learning & \% Acc \\ [0.5ex]
\hline
POOF~\cite{bergpoof} &  & \cmark & Sim-2-131 & POOF & SVM & 56.8 \\
Alignments~\cite{gaves2013fgvc} &  & \cmark & Trans-X-4 & Fisher & SVM & 62.7 \\
Symbiotic~\cite{chai2013symbiotic} &  & \cmark & Trans-1-1 & Fisher & SVM & 61.0 \\
DPD~\cite{zhang2013dpd} &  & \cmark & Trans-1-8 & KDES & SVM & 51.0 \\
Decaf~\cite{donahue2013decaf} &  & \cmark & Trans-1-8 & CNN & Logistic Regr. & 65.0\\
CUB~\cite{WahCUB_200_2011} &  &  & Trans-1-15 & BoW & SVM & 10.3\\
Visipedia~\cite{branson2014ignorant} &  &  & Trans-1-13 & Fisher & SVM & 56.5 \\
\textbf{Ours} & & & Sim-5-6 & CNN & SVM+CNN-FT & \textbf{75.7} \\
\hline
CUB Loc.~\cite{WahCUB_200_2011} & \cmark & \cmark & Trans-1-15 & BoW & SVM & 17.3\\
POOF Loc.~\cite{bergpoof} & \cmark & \cmark & Sim-2-131 & POOF & SVM & 73.3\\
\textbf{Ours Loc.} & \cmark & \cmark & Sim-5-6 & CNN & SVM+CNN-FT & \textbf{85.4} \\  [1ex]
\hline \\
\end{tabular}
\caption{Comparison to Related Work on CUB-200-2011: Our method significantly outperforms all earlier methods to our knowledge, both in terms of fully automatic classification accuracy (top grouping), and classification accuracy if part locations are provided at test time (bottom grouping).  We categorize each method according to 4 axes which we believe significantly affect performance: 1) \textit{Level of automation}, where column 2-3 indicate whether or not parts or object bounding boxes are assumed to be given at test time, 2) \textit{Part localization scheme} (column 4), using the naming scheme Transformation-X-Y, where Transformation indicates the image warping function used (see Table~\ref{tab:warp}), X indicates the number of keypoints/base-parts used to warp each region, and Y indicates the number of pose regions used, 3) \textit{Type of features} (column 5), and 4) \textit{Learning algorithm} (column 6), where CNN-FT is short for CNN fine-tuning.  }
\label{tab:related_comparison}
\end{table}

Table~\ref{tab:related_comparison} summarizes our main results and comparison to related work.  Our fully automatic approach achieves a classification accuracy of $75.7\%$, a $30\%$ reduction in error from the highest performing (to our knowledge) existing method~\cite{donahue2013decaf}.  We note that our method does not assume ground truth object bounding boxes are provided at test time (unlike many/most methods).  If we assume ground truth part locations are provided at test time, accuracy is boosted to $85.4\%$.  These results were obtained using prototype learning using a similarity warping function computed using 5 keypoints per region, CNN fine-tuning, and concatenating features from all layers of the CNN for each region.

We attempt to categorize each related method according to part localization scheme, features used, and learning method.  See the caption of Table~\ref{tab:related_comparison} for details.  The major factors that we believe explain performance trends and improvements are summarized below:
\begin{enumerate}[noitemsep,nolistsep]
\itemsep0em
 \item Choice of features caused the most significant jumps in performance.  The earliest methods that used bag-of-words features achieved performance in the $10-30\%$ range~\cite{WahCUB_200_2011,zhang2012pose}.  Recently methods that employed more modern features like POOF~\cite{bergpoof}, Fisher-encoded SIFT and color descriptors~\cite{perronnin2010improving}, and Kernel Descriptors (KDES)~\cite{bo2010kernel} significantly boosted performance into the $50-62\%$ range~\cite{bergpoof,chai2013symbiotic,gaves2013fgvc,zhang2013dpd,branson2014ignorant}.  CNN features~\cite{krizhevsky2012imagenet} have helped yield a second major jump in performance to $65-76\%$.
 \item Incorporating a stronger localization/alignment model is also important.  Among alignment models, a similarity transformation model fairly significantly outperformed a simpler translation-based model.  Using more keypoints to estimate warpings and learning pose regions yielded minor improvements in performance.
 \item When using CNN features, fine-tuning the weights of the network and extracting features from mid-level layers yielded substantial improvements in performance beyond what had been explored in~\cite{donahue2013decaf}.
\end{enumerate}
We support these conclusions by performing lesion studies in the next 3 sections. %with respect to features used (Section~\ref{sec:experiments_features}), part localization model (Section~\ref{sec:experiments_parts}), and CNN learning method (Section~\ref{sec:experiments_cnn}) 

\Subsection{Comparing Feature Representations}
\label{sec:experiments_features}

\begin{figure}[t]
\centering
  \subfigure[Feature Performance Comparison]{\label{fig:feature_comparison}
    \includegraphics[width=.46\textwidth]{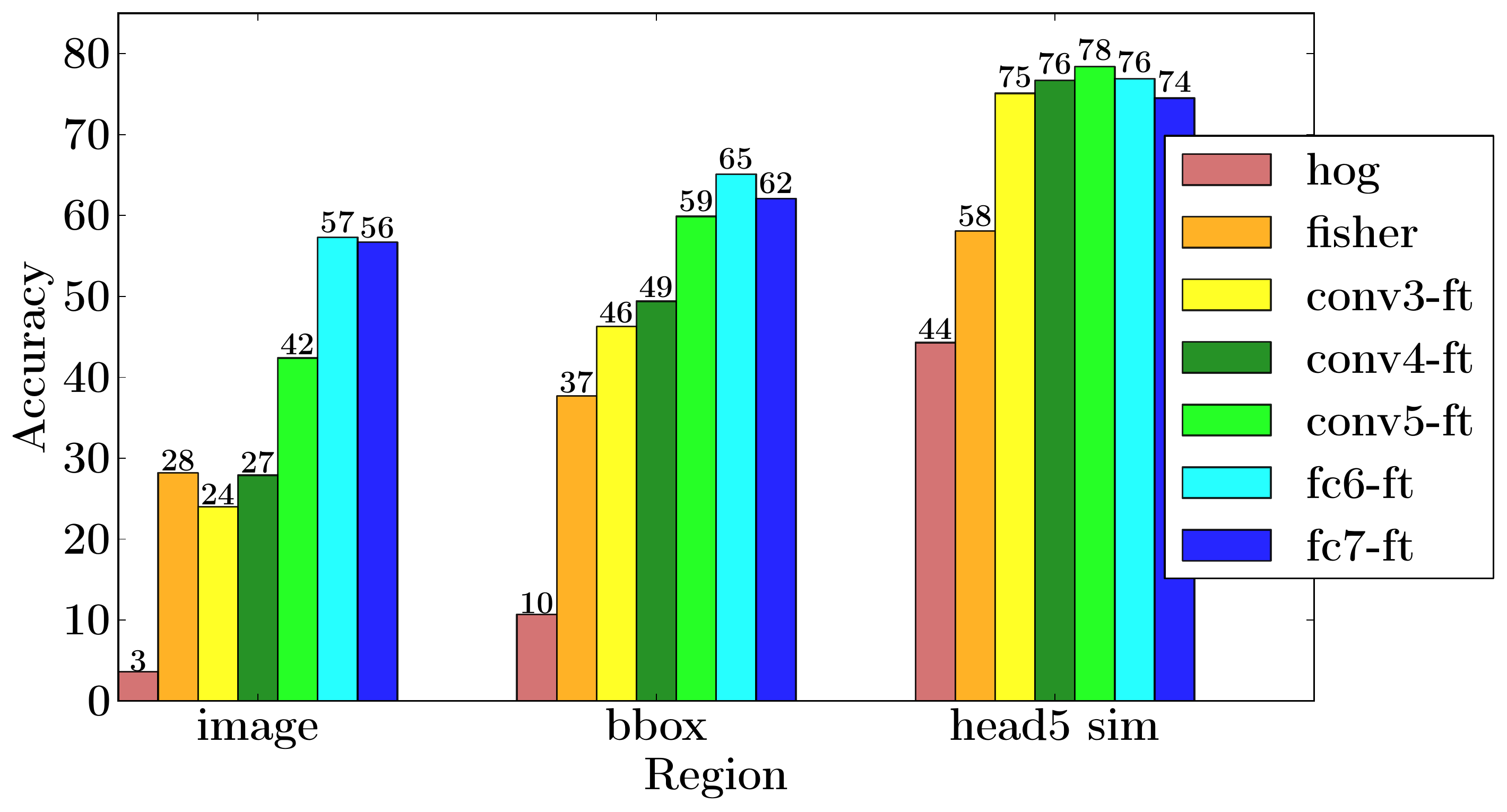}
   
  }
  \subfigure[Effect of CNN Layers For Different Regions]{\label{fig:region_comparison}
    \includegraphics[width=.50\textwidth]{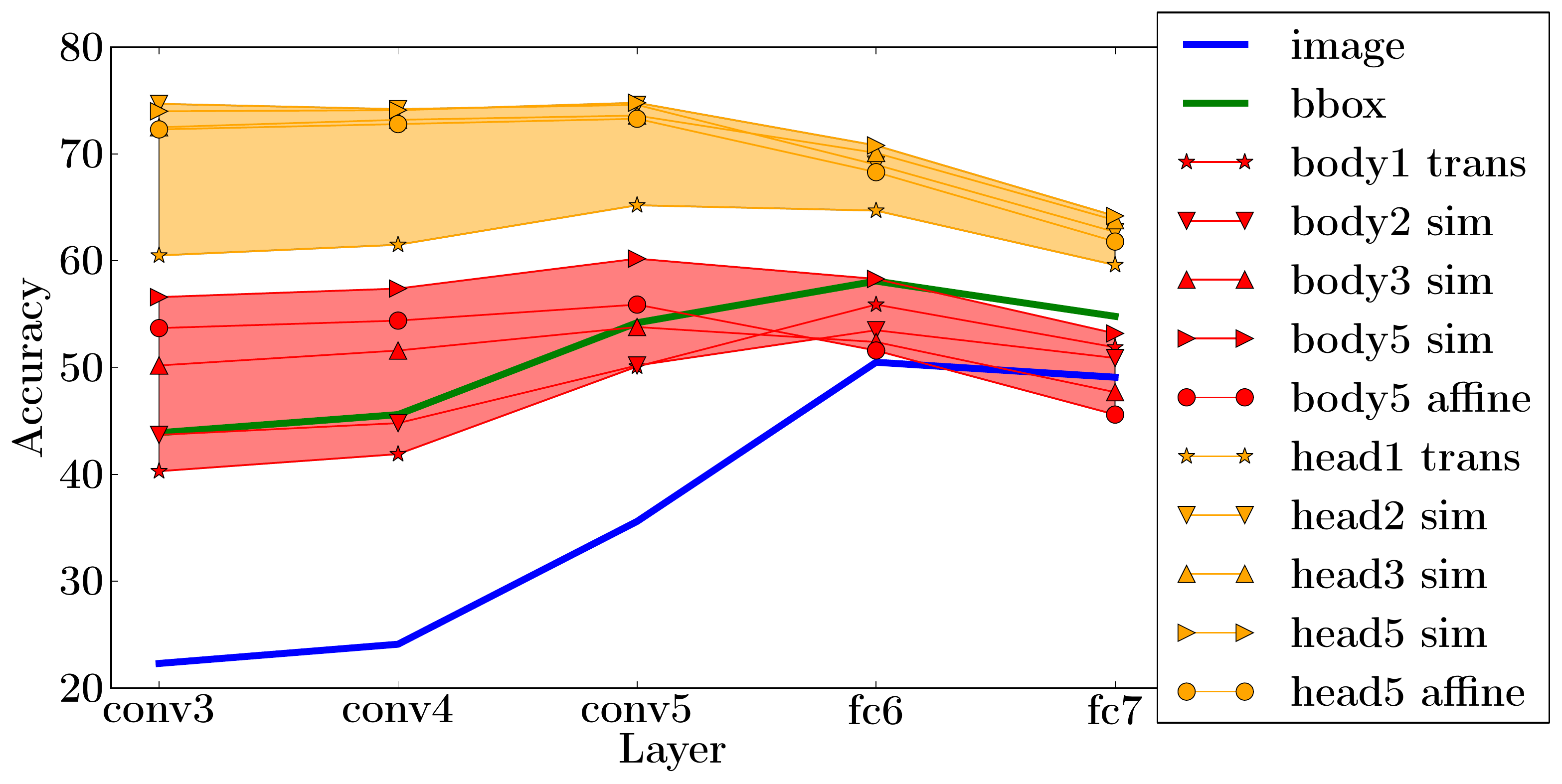}
  }
\Caption{{\bf Effect of features and region type on CUB-200-2011:}  \textbf{(a)}  CNN features significantly outperform HOG and Fisher features for all levels of alignment (image, bounding box, head). \textbf{(b)} Comparing classification performance for different CNN layers and regions if we assume ground truth part locations are known at test time,  we see that 1) features extracted from the head (yellow tube) significantly outperform other regions, 2) The later fully connected layers (fc6 \& fc7) significantly outperform earlier layers when a crude alignment model is used (image-level alignment), whereas convolutional layers (conv5) begin to dominate performance as we move to a stronger alignment model (from image $\to$ bbox $\to$ body $\to$ head), 3) Using a similarity warping model significantly outperforms a translation model (width of the red and yellow tubes), and slightly outperforms an affine model, 4) Using more points (from 1 to 5) to estimate the warping improves performance for the body, whereas 2 points is sufficient for the head. }
\label{fig:feature_comparison}
\end{figure}

We performed experiments to quantify the effect of different image features and their performance properties with different alignment models.  We compare CNN features from different layers to HOG~\cite{dalal2005histograms} and Fisher-encoded~\cite{perronnin2010improving} color and SIFT features while controlling for other aspects of our algorithms.  HOG is widely used as a good feature for localized models, whereas Fisher-encoded SIFT is widely used on CUB-200-2011 with state-of-the-art results~\cite{chai2013symbiotic,gaves2013fgvc,branson2014ignorant}. For HOG, we use the implementation/parameter settings of~\cite{felzenszwalb2010object} and induce a $16 \times 16 \times 31$ descriptor for each region type.  For Fisher features, we use the implementation and parameter settings from~\cite{branson2014ignorant}.  We summarize the results below:

\paragraphheader{CNN features significantly improve performance:} In Fig~\ref{fig:feature_comparison}, we see that CNN features significantly outperform other features for all levels of alignment, $57.3\%$ vs. $28.2\%$ for image-level features, and $78.4\%$ vs. $58.1\%$ for a similarity-aligned head.  HOG performs well only for aligned regions (the head), while  Fisher features perform fairly well across different levels of alignment.
\paragraphheader{Different layers of the CNN are appropriate for different alignment models:} In Fig.~\ref{fig:region_comparison}, we see that the later fully connected layers of the CNN (fc6 \& fc7) significantly outperform earlier layers when a crude alignment model is used ($57.3\%$ vs $42.4\%$ for image-level alignment), whereas convolutional layers (conv5) begin to dominate performance as we move to a stronger alignment model (from image $\to$ bbox $\to$ body $\to$ head).  

%\begin{figure}
%\includegraphics[width=1.0\textwidth]{figs/RegionExamples}
%\end{figure}

\Subsection{Comparing Part Localization Schemes}
\label{sec:experiments_parts}

We next perform experiments to quantify the effect of our pose normalization scheme, including the effect of the type of warping function used, a comparison of different methods of combining multiple pose regions, and the effect of imperfect part detection.  

\paragraphheader{A similarity alignment model works best:} In Fig.~\ref{fig:region_comparison}, we compare the effect of different choices of warping functions (translation, similarity, and affine) and the number of keypoints used to estimate them.  We see that a similarity warping model significantly outperforms a translation model and slightly outperforms an affine model (on the head region, $74.8\%$ for similarity vs. $65.2\%$ for translation vs. $73.3\%$ for affine).  Secondly, we see that using more points (from 1 to 5) to estimate the warping improves performance for the body, whereas 2 points is sufficient for the head.

\begin{figure}[t]
\centering
  \subfigure[Improvement From GT Parts]{\label{fig:ground_truth_improvements}
    \includegraphics[width=.31\textwidth]{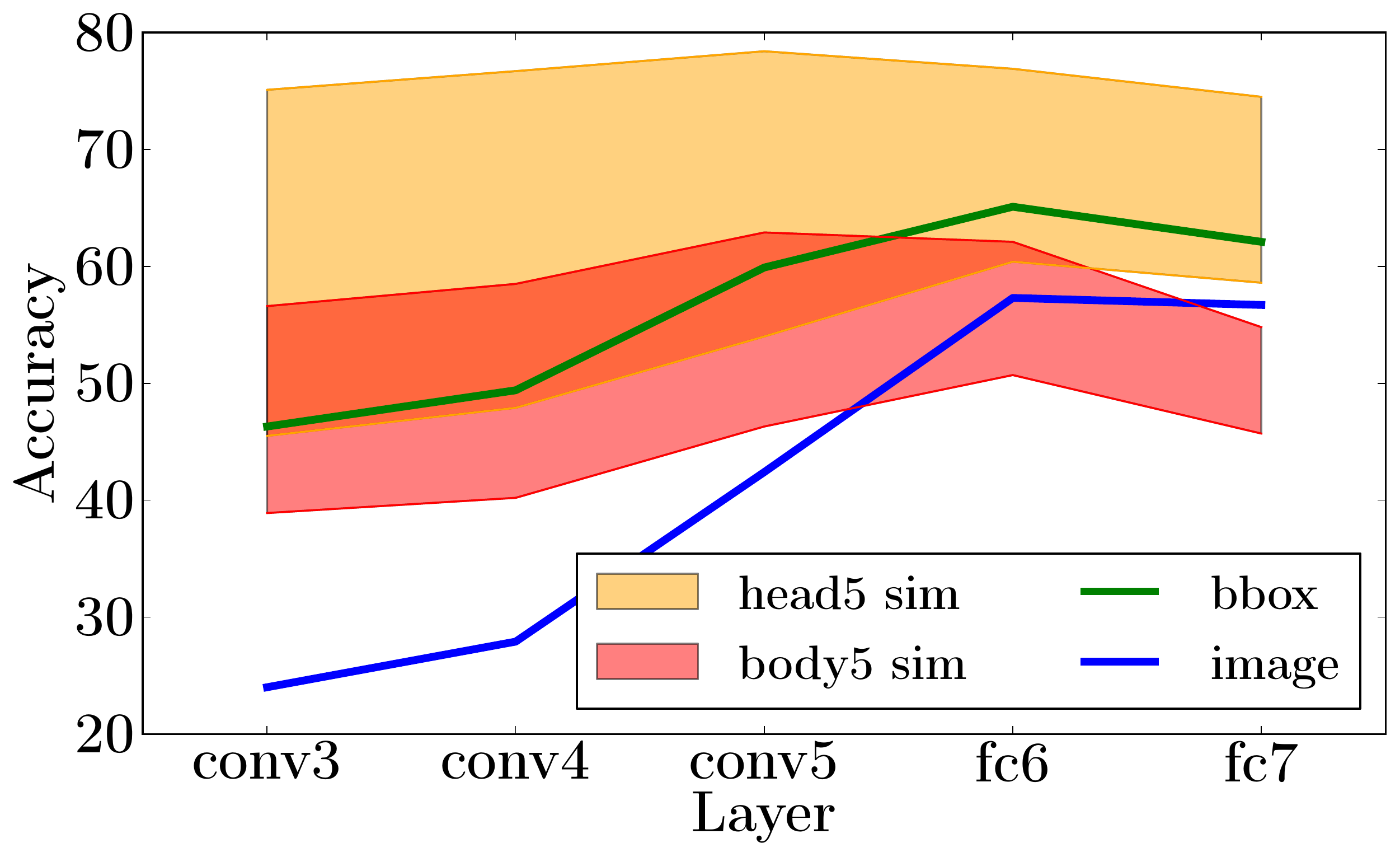}
  }
  \subfigure[Effect of Fine-Tuning, GT Parts]{\label{fig:fine_tuning_improvements_gt}
    \includegraphics[width=.31\textwidth]{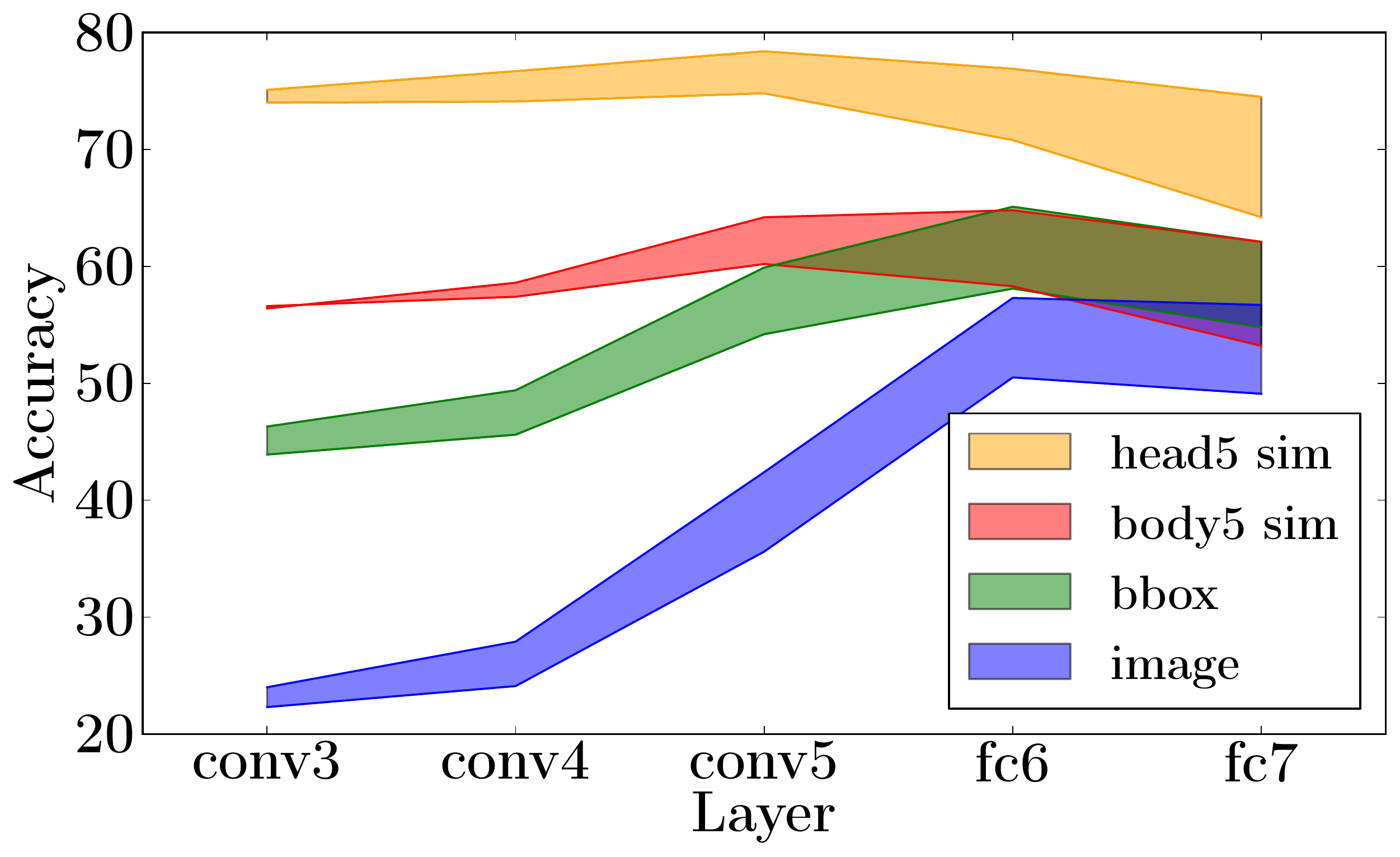}
  }
  \subfigure[Effect of Fine-Tuning, Pred Parts]{\label{fig:fine_tuning_improvements_pred}
    \includegraphics[width=.31\textwidth]{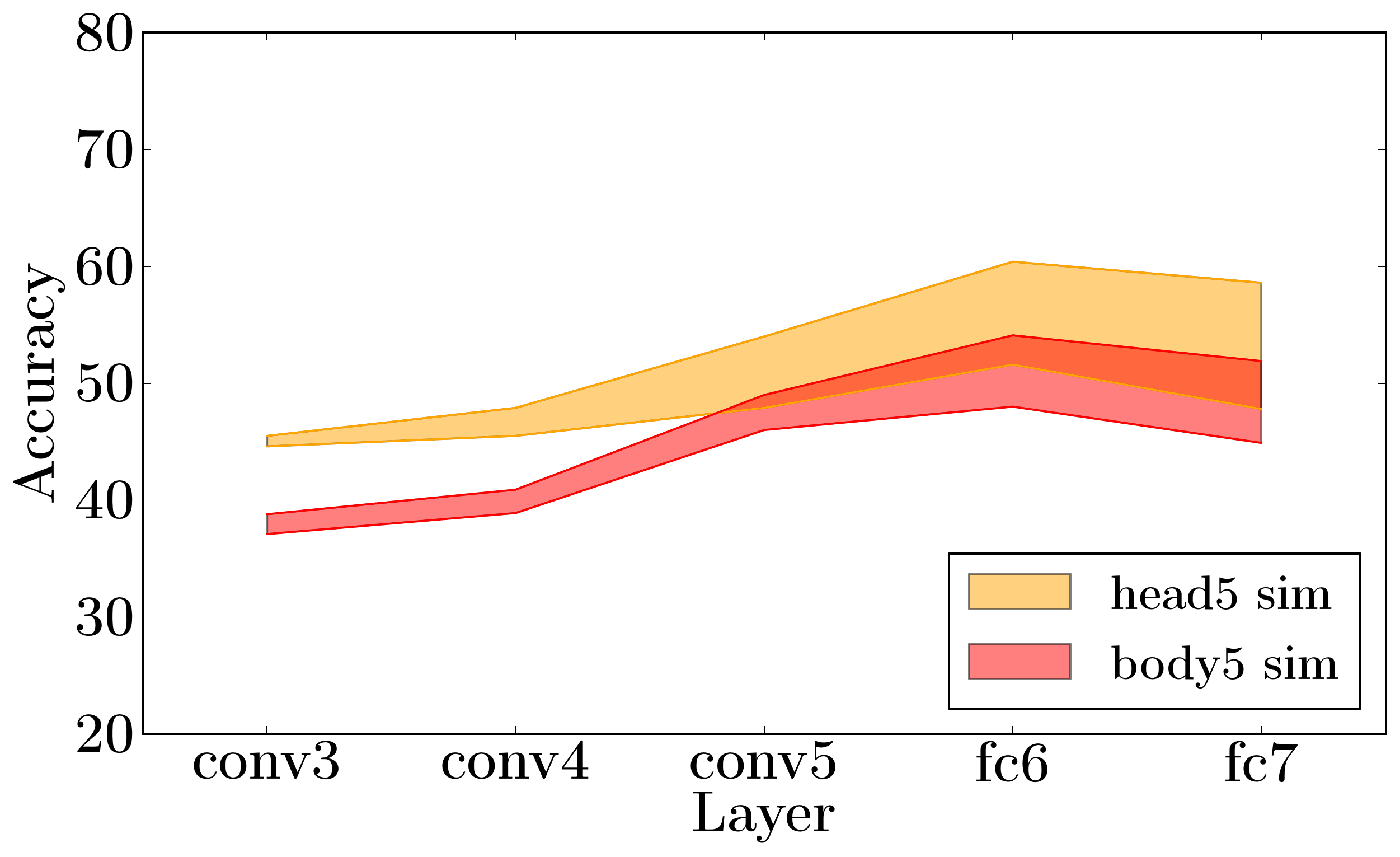}
  }
\Caption{{\bf Effect of fine-tuning and ground truth parts on CUB-200-2011:}  \textbf{(a)} If ground truth parts were available at test time or part detection could be improved, performance would be improved significantly (width of red/yellow tubes). \textbf{(b)} Fine-tuning significantly improves performance for all alignment levels (width of each tube).  Improvements occur for all CNN layers; however, the effect is largest for fully connected layers. \textbf{(c)} The same effect holds for automated part prediction.   }
\label{fig:improvements}
\end{figure}

\paragraphheader{Combining multiple regions improves performance:} In Table~\ref{tab:combine}, we compare different strategies for combining multiple pose regions.  We note that combining multiple regions improves performance over the best single region: $85.4\%$ vs. $78.4\%$ for the head.  We compare to several different baseline methods for inducing a multi-region feature space while keeping our feature implementation fixed.  The Proto-Learn method employs our pose learning scheme from Section~\ref{sec:pose_learn} using a similarity warping model and slightly outperforms other methods while being compact.  Rand-Pairs simulates the alignment method used by POOF~\cite{bergpoof}, where random pairs of keypoints induce similarity-aligned regions.  CUB-Keypoints simulates the method used by~\cite{branson2014ignorant} (among others), where each detected keypoint directly induces a surrounding pose region.  Head-Body represents a baseline of expert-defined regions, and concatenates hand-defined similarity-aligned head and body regions with image and bounding box features.

\squeeze
\squeeze
\begin{table}[ht]
\centering      
\footnotesize         
\begin{tabular}{ccccccccc}
\hline
 Head Body (2) & Proto-Learn (6) & Rand-Pairs (6) & Rand-Pairs (30) & CUB-Keypoints (13)\\  [0.5ex]            
83.7 & 85.4 & 83.2 & 84.1 & 79.6 \\    
\hline  
\\
\end{tabular}   
\caption{\textbf{Comparing Different stategies for combining multiple regions} when part locations are given at test time.  The number in parentheses indicates the number of regions used for each method.  }%Our method for learning regions automatically performs slightly better than other methods, while resulting in a compact feature space. }   
\label{tab:combine}
\end{table}
\squeeze
\squeeze

\paragraphheader{Imperfect part detection causes a significant but manageable drop in performance:} In Fig.~\ref{fig:ground_truth_improvements}, we visualize the drop in performance caused by using detected parts vs. ground truth parts for different regions, which ultimately results in a drop in performance from $85.4\%$ to $75.7\%$.  This is a sizeable drop in performance that we hope to reduce in future work by improving our part detection method; however, this gap is also surprisingly small, in large part due to the excellent performance of CNN features on image-level features.

\Subsection{Comparing CNN Learning Methods}
\label{sec:experiments_cnn}

In this section, we compare different strategies for learning the internal weights of the CNN.  

\paragraphheader{Fine-tuning CNN weights consistently improves performance:} In Fig.~\ref{fig:fine_tuning_improvements_gt}-\ref{fig:fine_tuning_improvements_pred}, we compare performance when using the pre-trained ImageNet model as a feature extractor vs. fine-tuning the ImageNet model on the CUB-200-2011 dataset (see Section~\ref{sec:cnn_train} for details).  We see that fine-tuning improves performance by $2-10\%$, and improvements occur for all region types (image, bounding box, head, body), all CNN layers, and both on predicted and ground truth parts.

\iffalse
\begin{table}[ht]
\centering
\footnotesize
\begin{tabular}{|r|cccc|}
\hline
Region & No Fine-Tuning & Fine-Tuning~\cite{girshick2013rich} & 2-Step Fine-Tuning & Train From Scratch\\
\hline
image  & 49.1 & 55.1 & 57.0 & 10.9\\
head sim & 74.8 & 76.9 & 78.6 & 54.7\\
\hline
\end{tabular}  
\vspace{2mm}
\caption{\textbf{Comparing CNN Training Methods: } Average classification accuracy over 5 random trials for the 4 methods described in Section~\ref{sec:cnn_train}.  Accuracies are shown for the fc7 image-level features and conv5 head-similarity features (the best layers for these two types of regions).}
\label{tab:ft}
\end{table}
\fi

\paragraphheader{ImageNet pre-training is essential:} The default CNN implementation was pre-trained on ImageNet and performance improvements come in part from this additional training data.  We tried training the same network structure from scratch on the CUB-200-2011 dataset over 5 trials with random initialization.  Performance was significantly worse, with $10.9\%$ and $54.7\%$ accuracy on image-level and similarity-aligned head regions, respectively (compared to $57.0\%$ and $78.6\%$ performance with pre-training) The problem relates to overfitting--the CNN model has 60 million learnable parameters~\cite{krizhevsky2012imagenet} and the CUB-200-2011 dataset has $<6000$ training images.  Learning converged to near zero training error for both fine-tuning and training from scratch. 

\paragraphheader{The two step fine-tuning method yields more reliable improvements:} Over 5 random trials, our proposed 2-step fine-tuning method improved average accuracy on both the image and head regions by about $2\%$ compared to the method used in~\cite{girshick2013rich} %while also resulting in faster convergence 
($57.0\%$ and $78.6\%$ compared to $55.1\%$ and $76.9\%$).

\Section{Conclusion}
\label{sec:conclusion}
In this paper, we reduced the error rate on CUB-200-2011 by $30\%$ compared to previous state-of-the-art methods, and analyzed which design decisions were most important to achieving good performance.  Our method is based on part detection and extracting CNN features from multiple pose-normalized regions.  Performance improvements resulted in large part from 1) using CNN features that were fine-tuned on CUB-200-2011 for each region, 2) using different CNN layers for different types of alignment levels, 3) using a similarity-based warping function that is estimated using larger numbers of detected keypoints.  We also introduced a novel method for learning a set of pose regions that explicitly minimizes pixel alignment error and works for complex pose warping functions.  In future work, we hope to apply our methods to other fine-grained datasets and explore customized CNN network structures and their training.
\Section{Acknowledgments}
\label{sec:acknoledments}
This work is supported by a Google Focused Research Award.

\bibliography{bmvc2014_birds}
\end{document}